\documentclass[letterpaper,10pt,conference]{ieeeconf}
\IEEEoverridecommandlockouts
\usepackage{algorithm}
\usepackage{algorithmic}
\usepackage{amsfonts}
\usepackage{amsmath}
\usepackage{amssymb}
\usepackage{array}
\usepackage{bm}
\usepackage{breakcites}
\usepackage{color}
\usepackage{comment}
\usepackage{float}
\usepackage{gensymb}
\usepackage{graphicx}
\usepackage{multirow}
\usepackage{pdfpages}
\usepackage{pgfplots}
\usepackage[caption=false,font=footnotesize,subrefformat=parens,labelformat=parens]{subfig}
\usepackage{textcomp}
\usepackage{tikz}
\usepackage{pgfplots}
\usepackage{wrapfig}
\usepackage{xcolor}
\usepackage{tabularx}
\usepackage{tikz}
\usetikzlibrary{spy}

\PassOptionsToPackage{hyphens}{url}\usepackage[breaklinks=true,hidelinks=true]{hyperref}

\graphicspath{{figures/}}

\definecolor{dargreen}{rgb}{0.0, 0.5, 0.0}

\title{Variable Rate Compression for Raw 3D Point Clouds}

\author{Md Ahmed Al Muzaddid and William J. Beksi 
\thanks{The authors are with the Department of Computer Science and 
        Engineering, University of Texas at Arlington, Arlington, TX, USA. 
        Emails: 
        mdahmedal.muzaddid@mavs.uta.edu,
        william.beksi@uta.edu.
        }
}

\begin{document}
\maketitle
\pagestyle{plain}

\begin{abstract} 
In this paper, we propose a novel variable rate deep compression architecture
that operates on {\em raw} 3D point cloud data. The majority of learning-based
point cloud compression methods work on a downsampled representation of the
data. Moreover, many existing techniques require training multiple networks for
different compression rates to generate consolidated point clouds of varying
quality. In contrast, our network is capable of explicitly processing point
clouds and generating a compressed description at a comprehensive range of
bitrates. 
Furthermore, our approach ensures that there is no loss of information as a
result of the voxelization process and the density of the point cloud does not
affect the encoder/decoder performance. An extensive experimental evaluation
shows that our model obtains state-of-the-art results, it is computationally
efficient, and it can work directly with point cloud data thus avoiding an
expensive voxelized representation.
\end{abstract}


\section{Introduction}
\label{sec:introduction}
Among various data modalities such as audio, image, text, etc., 3D data in the
form of point clouds constitutes a growing portion of the spectrum.  Sensors for
capturing 3D point cloud data such as light detection and ranging (LiDAR),
stereo, structured light, and time-of-flight (ToF) have become increasingly
popular and economical. In comparison to images, high-dimensional information
can be described in 3D with immunity to variations in color, illumination, and
scale. Besides point clouds, there are numerous ways to depict 3D data including
meshes, CAD models, and volumetric representations. However, when compared to
the other data representations, point clouds offer the advantage of providing a
simpler, denser, and closer-to-reality description \cite{cai2017rgb}.

A point cloud consists of an unordered set of points where each point has a
Cartesian coordinate value in Euclidean space. Additional attributes may
include the color (i.e., RGB value) and surface normal associated with each
point. Robotics is at the forefront of utilizing point clouds created by 3D
sensor technology. Portrayal of the environment provided by these sensors can
facilitate 3D learning-based tasks like object/scene reconstruction, motion
planning and navigation, object detection and classification, pose estimation,
grasp detection, etc. \cite{guo2020deep}. Other pertinent
applications where point clouds are widely used include augmented and virtual
reality, cultural heritage, geographical information systems, self-driving
vehicles, and many more \cite{sugimoto2017trends}.

3D sensors generate enormous amounts of point cloud data at high frame rates.
For example, a 3D point cloud with 0.7 million points per frame at 30 frames per
second needs a bandwidth of approximately 500 MB/s for video \cite{cao20193d}.
Working with uncompressed point cloud data can lead to congestion and delays in
communication networks \cite{beksi2014point}. Thus, efficient compression coding
technologies are indispensable for ensuring the compact storage and transmission
of such data. There exists an array point cloud compression techniques in the
literature which are primarily based on the use of octrees and 2D-projection.
Nevertheless, a number of deep learning-based point cloud compression methods
have emerged which offer performance improvements over traditional methods.

\begin{figure}[t]
\centering
\includegraphics[scale=0.90]{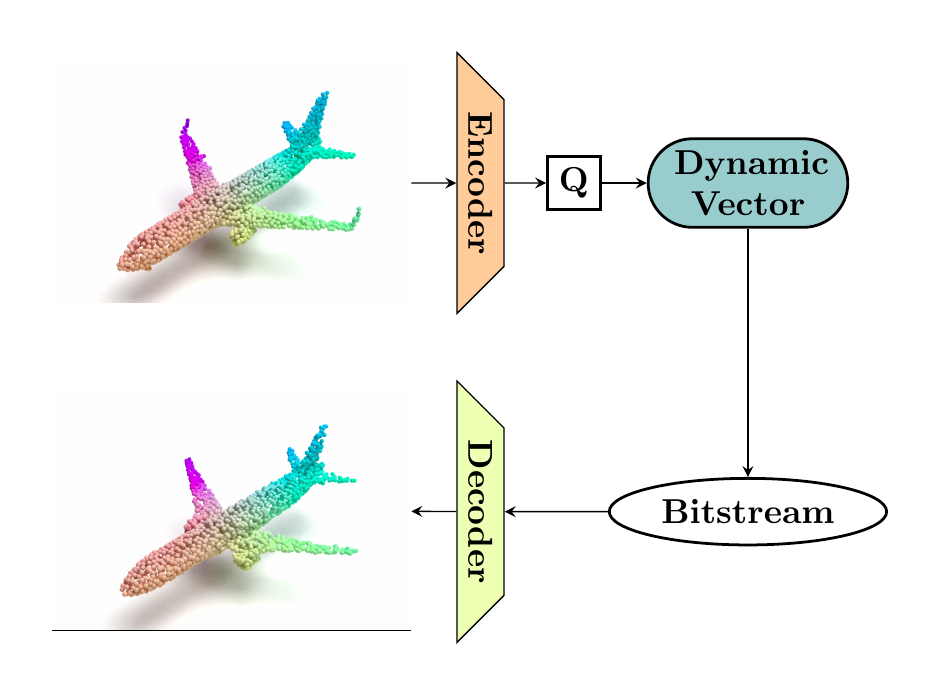}
\caption{3D point clouds produced by LiDAR, stereo, structured light, and ToF
sensors require a large amount of memory capacity. The transfer of point clouds 
to and from machines necessitates compression of the data. Our approach can
reconstruct dense point clouds from a highly-compressed representation, at
variable bitrates, using a single trained model.}
\label{fig:overview}
\end{figure} 

Even though most machine learning point cloud compression research has been
inspired by the success of learning-based image compression
\cite{patel2019survey}, there are fundamental differences between these two
domains. Unlike a traditional 2D image, the structure of a 3D point cloud is
irregular. Dealing with unordered 3D points is a significant dilemma. For
instance, proven techniques such as convolutional layers that extract features
in the image domain can not be plainly applied to raw point clouds. These 2D
and 3D convolution operators are suitable for grid-like data structures.
Therefore, point clouds are typically voxelized whereby points are represented
as a binary occupancy cell. Although this provides a representation suitable
for use with a convolutional layer, it also introduces a number of difficulties
(e.g., computing and memory overhead, loss of information, etc.). 


To address these limitations, we propose a model that can compress raw point
clouds without converting them into a voxelized representation,
Fig.~\ref{fig:overview}. There are two prominent benefits of our approach: (i)
there is no loss of information due to voxelization; (ii) the compression and
reconstruction performance is not affected by the underlying point cloud
density. At the same time, our technique reduces the unnecessary computational
burden induced by the processing of unoccupied voxels. In summary, our
contributions are the following.
\begin{itemize}
  \item We introduce a weighted entropy loss function and inference strategy to
  compress point clouds at different bitrates using a single trained model.
  \item Our architecture eliminates the need to train multiple models for a
  range of bitrates thus benefiting a diversity of applications.
  \item We establish raw 3D point cloud compression benchmarks for a variety of
  tasks on publicly available datasets.
\end{itemize}
Our source code is available at \cite{vrcpcd2022}.

The remainder of this paper is organized as follows. Relevant related literature
is discussed in Section~\ref{sec:related_work}. Our approach for variable rate
deep 3D point cloud compression is presented in
Section~\ref{sec:variable_rate_deep_compression}. The design and results of our
experiments are demonstrated and explained in
Section~\ref{sec:experimental_evaluation}. In
Section~\ref{sec:conclusion_and_future_work}, the paper is concluded and future
work is discussed.

\section{Related Work}
\label{sec:related_work}
\subsection{Tree Structure-Based Compression} 
Octrees provide an efficient way of representing point clouds through the
partitioning of space \cite{meagher1982geometric}. An octree-based
representation builds a tree data structure that starts from the root node,
where a node represents a 3D bounding box, and recursively subdivides the space
into eight child nodes until a desired depth is reached. 
Conventional octree-based coding methods \cite{huang2006octree,
schnabel2006octree,kammerl2012real,elseberg2013one,hornung2013octomap,
mekuria2016design} directly exploit the spatial correlations among points via
this data structure. 

Other types of tree-based representations, such as k-d trees trees,
\cite{bentley1975multidimensional} and spanning trees
\cite{cheriton1976finding}, can achieve analogous results for lossless coding
\cite{gandoin2002progressive,gumhold2005predictive,merry2006compression}. The
Moving Picture Experts Group (MPEG) has published a geometry-based point cloud
compression standard (MPEG V-PCC) based on an octree data structure
\cite{zakharchenko2018algorithm,schwarz2018emerging}. In addition, the Point
Cloud Library \cite{rusu20113d} provides octree-based implementations for point
cloud compression.  

\subsection{Projection-Based Compression} 
Projection-based approaches follow the principle of converting a 3D point cloud
into 2D images by projection or mapping \cite{ochotta2004compression,
merkle2007multi,golla2015real,he2017best,sun2019novel,tu2019point}. Usually,
the output of the projection step is represented as a set of depth images based
on multiple dissimilar viewpoints or canonical planes. This provides a regular
grid structure of the points, which are further partitioned into a set of
patches or point clusters. Subsequent steps compress the patches using
well-known image/video codecs such as JPEG \cite{wallace1992jpeg} and HEVC
\cite{sullivan2012overview}. The main challenge with this technique is to find
an optimum set of transformations from the 3D point set to the corresponding 2D
representation. A combination of octree and projection-based methods
\cite{ainala2016improved} is also possible as well. 

\subsection{Learning-Based Compression} 
Learning-based methods typically exploit the same transform-based coding
principles used in conventional lossy image coding such as JPEG. Most of the
learning-based models search for an optimum nonlinear transformation in terms
of rate distortion trade-off during training time by using an appropriate
differentiable objective function. For instance, the estimated entropy of a
quantized latent vector can be used along with the reconstruction loss in an
objective function \cite{wang2019learned,guarda2020deep}. Within the image
domain, convolutional neural networks (CNNs) are used to extract the
translation-invariant features that encapsulate spatial correlations among
neighboring pixels. Inspired by the success of CNN layers
\cite{mentzer2018conditional}, complementary approaches have been adopted in
learning-based point cloud compression \cite{guarda2019deep,guarda2019point,
quach2019learning,quach2020improved,wiesmann2021deep}.

After converting a raw point cloud into a regular voxel grid, autoencoder-based
machine learning models may be used to extract a latent representation which is
then quantized. The quantized latent representation is further processed by a
decoder to reconstruct the point cloud. The whole model can then be trained
end-to-end to optimize both reconstruction loss and entropy loss. Due to the
non-differentiable nature of the quantization step, special treatment is needed
inside a conventional autoencoder structure. For example,
\cite{agustsson2017soft} carried out nearest neighbor assignments to compute the
quantized value of the latent vector which relied on (differentiable) soft
quantization. Motivated by theoretical links to dithering, \cite{balle2016end}
proposed to replace quantization by additive uniform noise.
\cite{theis2017lossy} used a stochastic rounding operation in the forward pass,
while in the backward pass the derivative is replaced with the derivative of the
expectation.

\section{Variable Rate Deep Compression}
\label{sec:variable_rate_deep_compression}
\begin{figure*}
\centering
\includegraphics[scale=0.37]{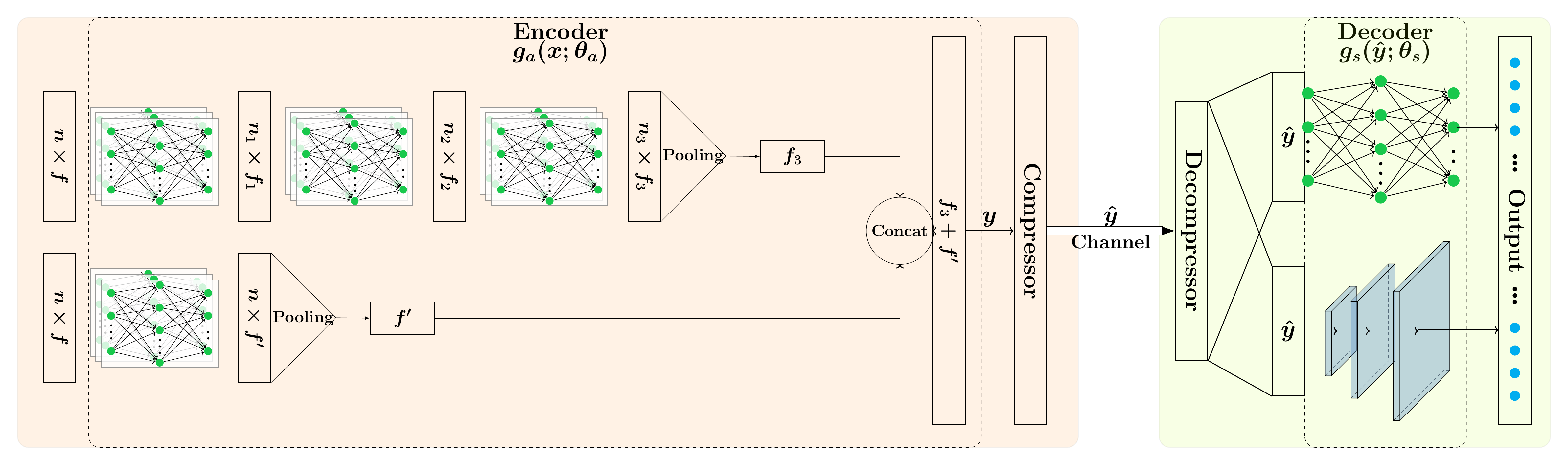}
\caption{An overview of our variable rate deep compression model for
compressing raw 3D point clouds. The input to the model is an $n \times f$
matrix, where $n$ is the number of sample points and $f$ is the number of
features per point. The top and bottom branches of the encoder correspond to
the local and global feature extractors, respectively. At the first level in
the local feature extractor, the set abstraction (SA) layer transforms the per
point input features into combined features of size $n_1 \times f_1$. These
combined features are then processed by the next level SA layer. All of the
features are combined into a single feature vector of size $f_3$ at the last
level SA layer. The bottom branch converts per point input features of size $n
\times f$ into transformed features of size $n \times f^\prime$. Next, a
pooling operation combines the features into a single feature vector of size
$f^{\prime}$. Lastly, the two feature vectors of size $f_3$ and $f^{\prime}$
are concatenated and processed by the compressor. The decoder first
decompresses the compressed vector into $\hat{y}$ which is then processed by
both of the decoding branches (top and bottom) to produce approximately $n$
points in total.}
\label{fig:model_architecture}
\end{figure*}  

Our unified architecture directly takes raw point clouds as input,
Fig.~\ref{fig:model_architecture}. A PointNet-based \cite{qi2017pointnet}
feature extractor is utilized for processing point cloud data as follows. The
encoder transforms a point cloud $x$ using a parametric analysis transform,
$g_a(x;\theta_a)$, into a latent representation $y$. The latent representation
is then quantized to produce $\hat{y}$. On the receiving side, the decoder
recovers $\hat{y}$ from the compressed signal. A synthesis transformer function,
$g_s(\hat{y};\theta_s)$, uses $\hat{y}$ to recover the reconstructed point cloud
$\tilde{x}$. $\theta_a$ and $\theta_s$ are the weight parameters of the encoder
and decoder, respectively.

Our goal is to derive uncorrelated global and local features from the point
cloud input. To do this, we propose a deep learning-based approach to extract
structural features from 3D data. Transforms (e.g., discrete cosine transform
(DCT) \cite{ahmed1974discrete}) have been used for decades to represent 2D image
and video data in a more condensed format. Although we employ a feature
extractor that encodes low-level features into compact vector representations,
floating-point vectors are not suited for sending and receiving over a
transmission channel in an efficient way. This is due to the fact that precise
floating-point data requires a high bitrate to transmit. 

To reduce the bitrate, we further quantize the embedded vector using an adaptive
quantizer coupled with a probability density model. The probability density
model is applied to estimate the bitrate required to compress the quantized
vector representation. A fully-connected decoder is used as a synthesis
transformer to decode the quantized latent features into the reconstructed point
cloud. The reconstructed point cloud is then compared against the input point
cloud and the reconstruction loss is calculated. The whole model is trained
end-to-end to simultaneously optimize reconstruction loss and bitrate. In the
following subsections, we describe each component of the model.

\subsection{3D Transform} 
Within the image domain, transformations are used to express data in a
consolidated form. Learned 2D transforms have demonstrated improved coding
performance in image compression \cite{balle2018variational,
mentzer2018conditional,choi2019variable}. A learned transform usually performs
better than a traditional transform such as the DCT. This is due to the ability
of the learned transform to allocate more or less bits for contrasting parts of
an image based on the rate distortion constraints it is trained on. For point
cloud data, a transform is typically learned via a convolutional autoencoder
which takes a voxelized occupancy grid as input. 

Instead of using a convolution-based feature extractor, we opt for a
fully-connected layer-based feature extractor. Concretely, we use two parallel
branches in our encoder module. One branch handles the global structure of the
point cloud and the other branch focuses on local patterns along the disparate
levels of granularity. In the global feature extractor branch, we apply a
multilayer perceptron network to every point. The network weights are shared
among all the points. A global feature vector is then derived by applying a
symmetric function on the transformed features.

In the local feature extractor branch, features are aggregated in a hierarchical
fashion to capture the geometric structure of the neighboring point cloud
distribution. The points are sampled based on an iterative farthest point
strategy \cite{qi2017pointnet++} and used as keypoints to extract nearby local
structure. Similar to a CNN, sampled features only focus on a small neighborhood
within the initial layers. At the higher layers, features are captured at larger
scales. This is comparable to the increasing receptive field size within the
deeper layers of a CNN. Both feature vectors from the local and global branch
are then concatenated and passed through fully-connected layers which combine
them into a single latent vector $y$. 

\subsection{Quantization}
After obtaining the latent representation, we perform a quantization step at
compression time based on the desired target quality or rate.  During training,
to ensure differentiability we model the discretization using additive uniform
noise \cite{balle2018variational}. More formally, we define
\begin{equation}
  q(\tilde{y}\,|\,x,\theta_s) = \prod_{i}U\left(\tilde{y}_i\,|\,y_i-\frac{1}{2},y_i+\frac{1}{2}\right),
\end{equation}
where $\tilde{y}$ is the noisy approximation, $U$ denotes a uniform
distribution centered on $y_i$, and $i \in [1,\ldots,l]$ where $l$ is the
length of $y$. During compression, the elements of $y$ are quantized by
rounding to the nearest integer.

\subsection{Rate-Distortion Modeling} 
Following quantization, the latent vector becomes discrete-valued. Consequently,
it can be compressed losslessly using entropy coding techniques such as
arithmetic coding \cite{rissanen1981universal}. The compressed code can then be
transmitted as a sequence of bits. On the receiving end, the decoder recovers
the 3D coordinates from the compressed latent vector to reconstruct the original
point cloud. Given an approximate distribution $P_{\hat{y}}$ of the quantized
vector $\hat{y}$, we calculate the bitrate as 
\begin{equation} 
  R = E_x[-\log P_{\hat{y}}(Q(g_a(x;\theta_a)))], 
\end{equation}
where $Q$ represents the quantization function. The distortion is the expected
difference between the reconstructed point cloud $\tilde{x}$ and the inceptive
point cloud $x$ as measured by a suitable distance metric. To control the rate
distortion trade-off we use a traditional formulation in which an equilibrium is
found between the code length and the distortion. Formally, we minimize the
combined loss, $L = D + \lambda R$, where $D$ is the distortion and $\lambda$
acts as a hyperparameter that balances the two competing objectives.

\subsection{Decoder}
Given a vector decompressed by the decompressor, the decoder module regresses
the 3D coordinate values of the point cloud. To support variable bitrates, we
use a weighted entropy loss function. Through the use of variable weights for
individual vector elements, we gradually reduce the importance of the elements
occurring at positions later on in the compressed vector. More precisely, 
\begin{equation} 
  R_{weighted} = E_x\left[\sum_{i} -\omega_i \log P(\hat{y}_i)\right],
  \label{eq:weighted_entropy}
\end{equation}
where 
\begin{equation} 
  \omega_i = a e^{-b\times i}
  \label{eqn:weight_decay}
\end{equation} 
is an exponential decaying function for assigning the $i$-th weight to each
element in the compressed vector. 

With this weighted entropy loss the overall loss function becomes $L = D +
\lambda R_{weighted}$. During the reconstruction phase, elements appearing
towards the end of the vector contribute less to the overall reconstruction. As
a result, we can ignore the latter part of the vector if we have a tight budget
on the bitrate. This allows us to train a single model that produces variable
bitrates. We show the relation between the effective bitrate versus the
reconstruction loss through a broad experimental assessment
(Section~\ref{sec:experimental_evaluation}).
 
\section{Experimental Evaluation} 
\label{sec:experimental_evaluation} 

\subsection{Datasets, Training, and Runtime Details} 
We make use of the ShapeNet \cite{chang2015shapenet} and ModelNet40
\cite{wu20153d} datasets for our experimental evaluation. ShapeNet is composed
of 16,881 objects from 16 classes and 50 parts in total. In addition, the
category label for each object is provided. Each point cloud contains 2,048
points uniformly sampled from the surface of an object. The sampled point clouds
are normalized with zero mean and scaled to a unit sphere. ModelNet40 consists
of 12,311 CAD-generated objects (e.g., airplane, car, plant, lamp, etc.) from 40
categories. Within ModelNet40, 9,843 objects are used for training while the
rest (2,468) are reserved for testing.  

Our model was trained in batches where each sample consists of $n=2048$ points.
Only the $x,y,z$ coordinates are considered as input features (i.e., $f= 3$)
akin to \cite{qi2017pointnet}. For the local feature extractor branch, three SA
layers were used. The number of sampling points in each SA layer are $n_1=512$,
$n_2=128$, $n_3=1$, and the number of features are $f_1=256$, $f_2=512$,
$f_3=512$. In the global feature extractor branch the number of features is
$f^\prime=512$. On the decoder side, we used a small fully-connected neural
network of size $512 \times 512 \times 1024$. Max pooling was utilized for all
pooling operations.

Using the Adam \cite{kingma2014adam} optimizer, our network was trained for 300
iterations with an initial learning rate 0.0001, a batch size of 16, momentum
value of 0.9, and momentum2 value of 0.999. Training took place on a CentOS
7.6.1810 machine using an Intel Xeon E5-2620 2.10 GHz CPU, 132 GB of memory, and
an NVIDIA GeForce GTX 1080 Ti GPU. It took approximately 23 hours to train our
model. All experimental tests were performed using a Ubuntu 18.04 machine with
an Intel Core i7-8700 3.20 GHz CPU, 32 GB of memory, and an NVIDIA Quadro P4000
GPU. Table~\ref{tab:elasped_wall-clock_time} displays the elapsed wall-clock
times, per batch size, for 3D point cloud compression and decompression. Note
that due to the parallel processing capability of the GPU, compression and
decompression in larger batch sizes takes less time per point cloud.

\newcolumntype{C}{>{\centering\arraybackslash}p{0.2\linewidth}}
\begin{table}
\begin{center}
\begin{tabular}{|C|C|C|C|} 
  \hline
  \multirow{2}{*}{\textbf{Task}} & \multicolumn{3}{|c|}{\textbf{Time} (msec)}\\
  \cline{2-4}&
   Batch size of 32 & Batch size of 16 & Batch size of \newline 1\\
    \hline
    Compression   & $8.50 \pm0.10$ & $8.60 \pm0.10$ & $15.20 \pm0.40$\\
    Decompression & $0.40 \pm0.01$ & $0.50 \pm0.02$ & $2.30 \pm0.10$\\
  \hline
\end{tabular}
\caption{Elapsed wall-clock times for 3D point cloud compression and
decompression tasks averaged over ten trial runs.}
\label{tab:elasped_wall-clock_time}
\end{center}
\end{table}

\newcolumntype{s}{>{\hsize=.7\hsize}X}
\begin{table*}
\begin{tabularx}{\textwidth}{X|sss|sss|sss} \hline
\multirow{2}{*}{Category} & \multicolumn{3}{c|}{CD ($\times10^{-2}$) \textcolor{dargreen}{ $\downarrow$}} & \multicolumn{3}{c|}{EMD ($\times10^{-2}$) \textcolor{dargreen}{ $\downarrow$}} &\multicolumn{3}{c}{\textit{F}-score \textcolor{dargreen}{$\uparrow$} }\\ \cline{2-10}
          &1.69 bpp & 0.35 bpp & 0.03 bpp &1.69 bpp & 0.35 bpp & 0.03 bpp &1.69 bpp & 0.35 bpp & 0.03 bpp  \\ 
  \hline
  Airplane &0.063 &0.068 &0.074 &2.041 &2.361 &2.170 &0.995 &0.995 &0.992\\ 
  Bag & 0.343 & 0.362 & 0.392 & 2.733 & 3.346 & 2.439 & 0.788 &0.764 &0.744\\  
  Cap & 0.550 & 0.600  & 0.595 & 2.386 & 2.298 & 3.151 & 0.652 & 0.591 &0.623\\ 
  Car & 0.230 & 0.255 & 0.257 & 3.032 & 2.394 & 2.837 & 0.891 &0.884 &0.870\\ 
  \hline
  Chair & 0.202 & 0.211 & 0.273 & 2.575 & 2.673 & 3.016 & 0.919 &0.913 &0.879\\  
  Earphone & 0.463 & 0.457  & 0.656 & 3.237 & 2.192 & 2.822 & 0.762 &0.784 &0.735\\  
  Guitar & 0.055 & 0.056 & 0.067 & 1.846 & 1.815 & 2.406 & 0.990 &0.990 &0.986\\ 
  Knife & 0.072 & 0.070 & 0.086 & 1.408 & 1.901 & 1.963 & 0.987 &0.989 &0.983\\
  \hline
  Lamp & 0.321 & 0.349  & 0.389 & 2.274 & 3.054 & 3.091 & 0.845 &0.826 &0.807\\  
  Laptop & 0.127 & 0.132 & 0.165 & 2.313 & 2.535 & 2.566 &0.987 &0.985 &0.977\\ 
  Motorbike & 0.256 & 0.280 & 0.292 & 2.501 & 3.085 & 2.444 & 0.868 &0.857 &0.848\\  
  Mug & 0.369 & 0.407  & 0.453 & 2.446 & 2.305 & 2.239 & 0.726 &0.694 &0.623\\ 
  \hline
  Pistol & 0.110 & 0.128  & 0.138 & 2.611 & 2.276 & 2.206 & 0.974 &0.970 &0.963\\  
  Rocket & 0.162 & 0.156 & 0.173 & 2.072 & 2.481 & 2.314 &0.917 &0.921 &0.911\\ 
  Skateboard & 0.174 & 0.197 & 0.203 & 2.050 & 2.477 & 2.648 &0.941 &0.933 &0.930\\  
  Table & 0.179 & 0.195  & 0.230 & 2.039 & 2.475 & 2.690 &0.934 &0.923 &0.903\\ 
  \hline
  Mean & 0.230 & 0.245 & 0.278 &2.347 &2.479 &2.562 &0.886 &0.876 &0.861\\
  \hline
\end{tabularx}
\caption{Reconstruction loss via the chamfer distance (CD) and earth mover's
distance (EMD) across a range of bits per point (bpp) for each object category
in the ShapeNet \cite{chang2015shapenet} dataset. The F-score was calculated at
a distance threshold of $d=0.05$.}
\label{tab:loss_per_category}
\end{table*}

\begin{figure*}
\centering
\subfloat{
  \begin{tikzpicture}[spy using outlines={circle,red,magnification=3,size=1.5cm, connect spies}]
    \node {\includegraphics[width=.16\textwidth]{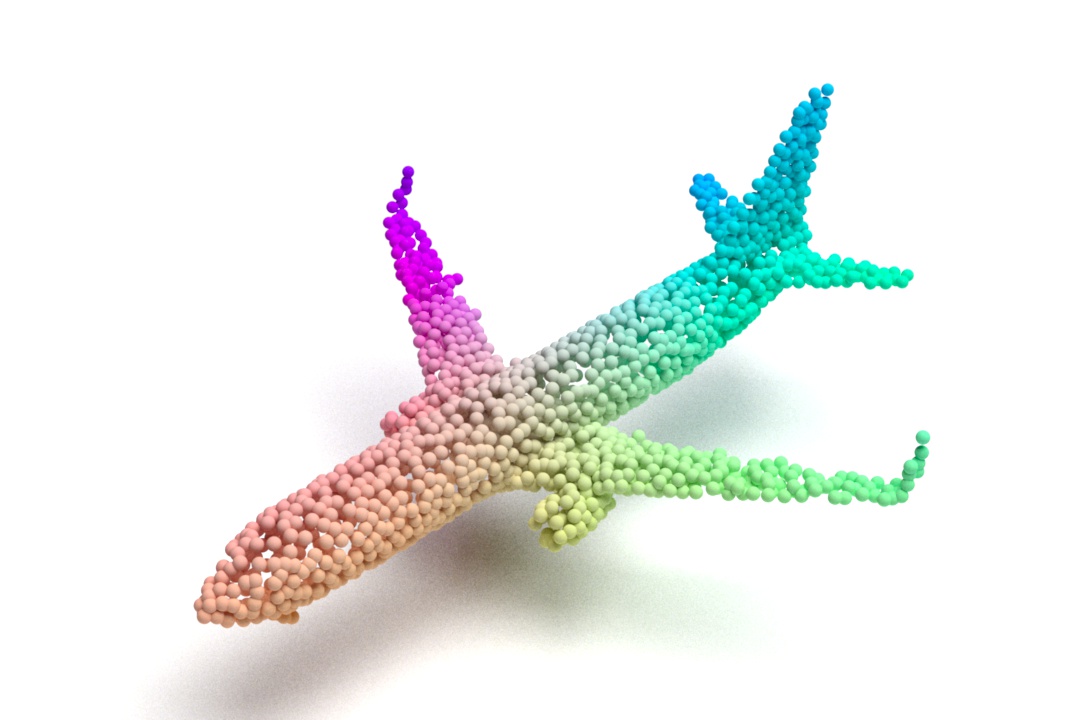}};
    \spy on (.05,-0.4) in node [left] at (-.3,.7);
  \end{tikzpicture}
} 
\subfloat{ 
  \begin{tikzpicture}[spy using outlines={circle,red,magnification=3,size=1.5cm, connect spies}]
    \node {\includegraphics[width=.16\textwidth]{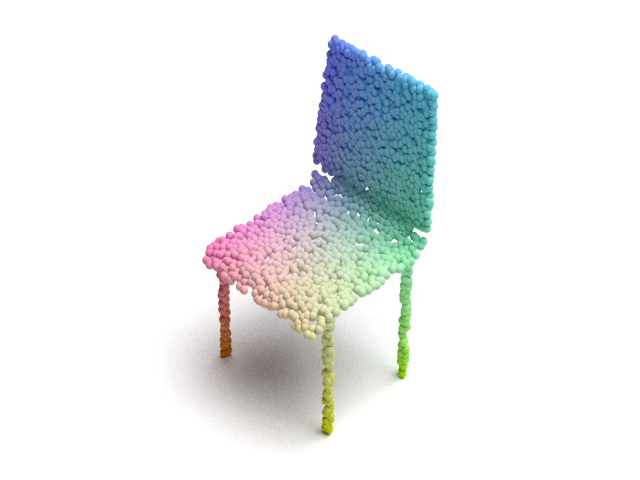}};
    \spy on (.05,-0.4) in node [left] at (-.3,.7);
  \end{tikzpicture}
} 
\subfloat{ 
  \begin{tikzpicture}[spy using outlines={circle,red,magnification=3,size=1.5cm, connect spies}]
    \node {\includegraphics[width=.16\textwidth]{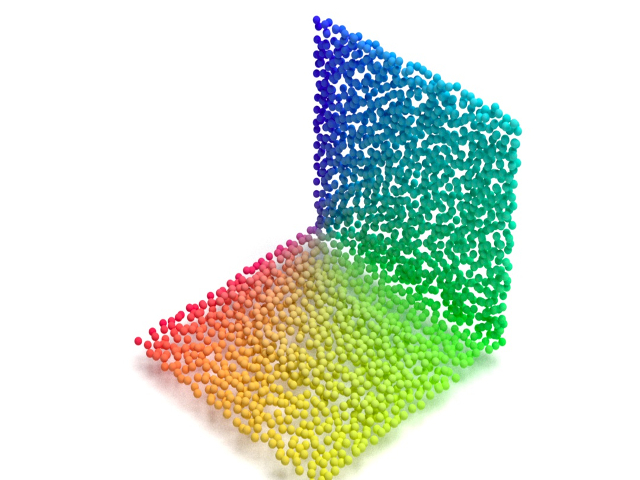}};
    \spy on (-.73,-0.4) in node [left] at (-.3,.7);
  \end{tikzpicture}
}
\subfloat{ 
  \begin{tikzpicture}[spy using outlines={circle,red,magnification=3,size=1.5cm, connect spies}]
    \node {\includegraphics[width=.16\textwidth]{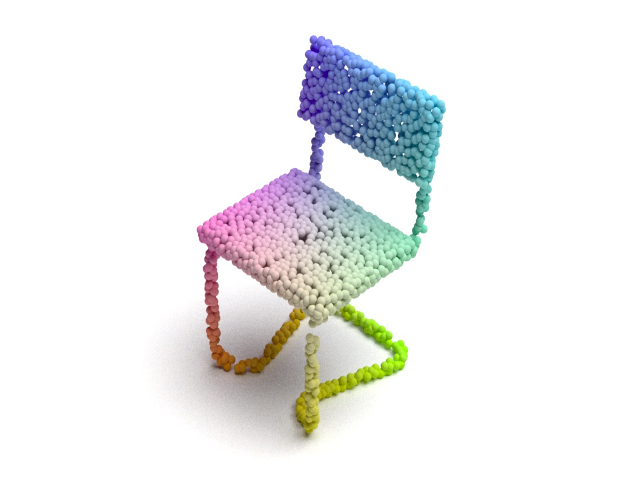}};
    \spy on (.0,-0.75) in node [left] at (-.3,.7);
  \end{tikzpicture}
} 
\subfloat{ 
  \begin{tikzpicture}[spy using outlines={circle,red,magnification=3,size=1.5cm, connect spies}]
    \node {\includegraphics[width=.16\textwidth]{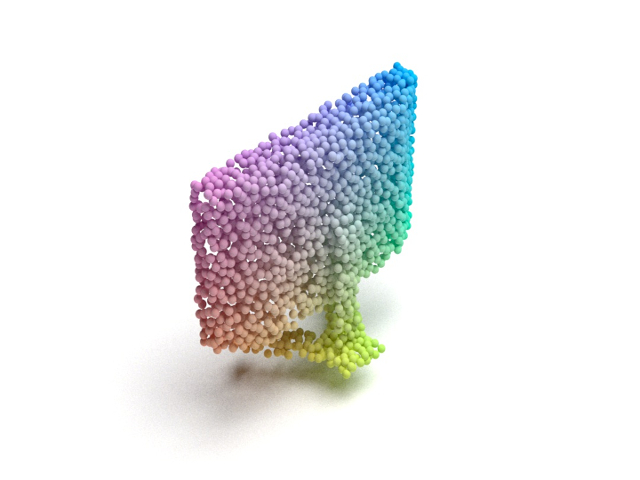}};
    \spy on (.1,-0.5) in node [left] at (-.3,.7);
  \end{tikzpicture}
} 
\\
\subfloat{ 

\begin{tikzpicture}[spy using outlines={circle,red,magnification=3,size=1.5cm, connect spies}]
    \node {\includegraphics[width=.16\textwidth]{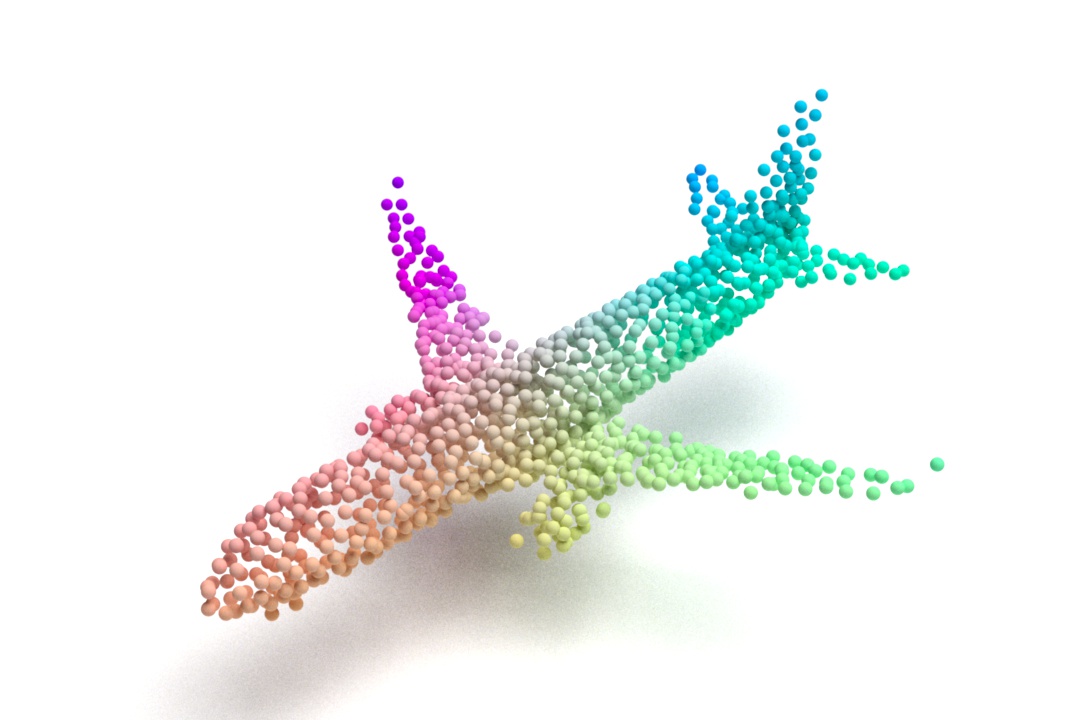}};
    \spy on (.05,-0.4) in node [left] at (-.3,.7);
\end{tikzpicture}
}
\subfloat{ 
   \begin{tikzpicture}[spy using outlines={circle,red,magnification=3,size=1.5cm, connect spies}]
    \node {\includegraphics[width=.16\textwidth]{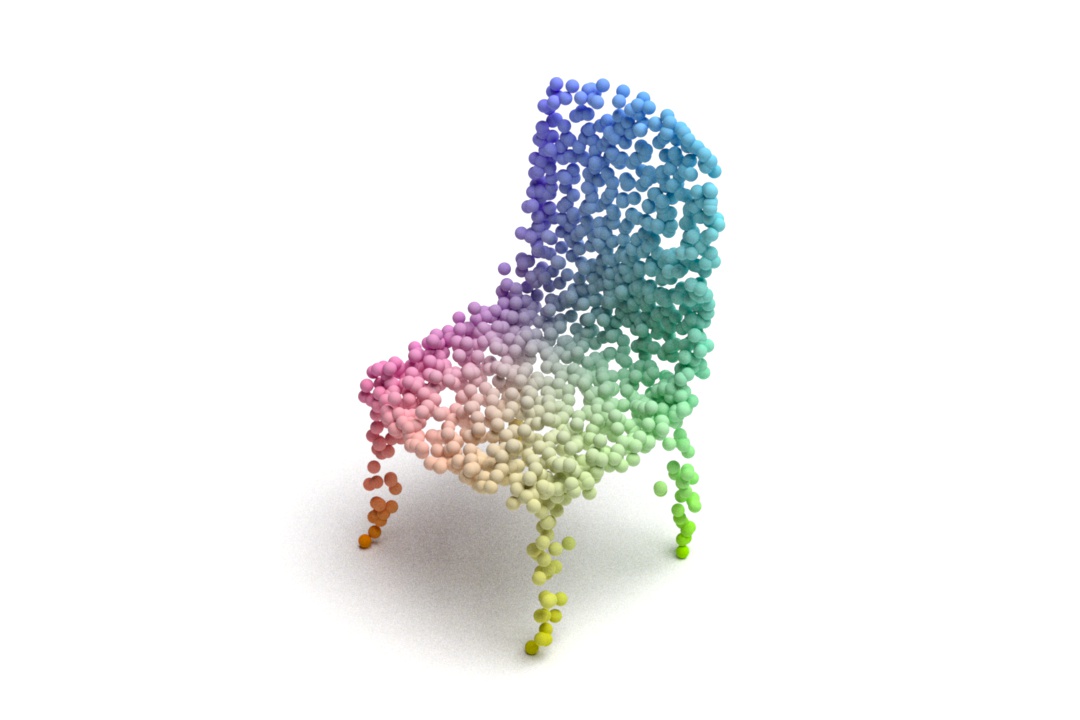}};
    \spy on (.07,-0.4) in node [left] at (-.3,.7);
  \end{tikzpicture}
}
\subfloat{ 
  \begin{tikzpicture}[spy using outlines={circle,red,magnification=3,size=1.5cm, connect spies}]
    \node {\includegraphics[width=.16\textwidth]{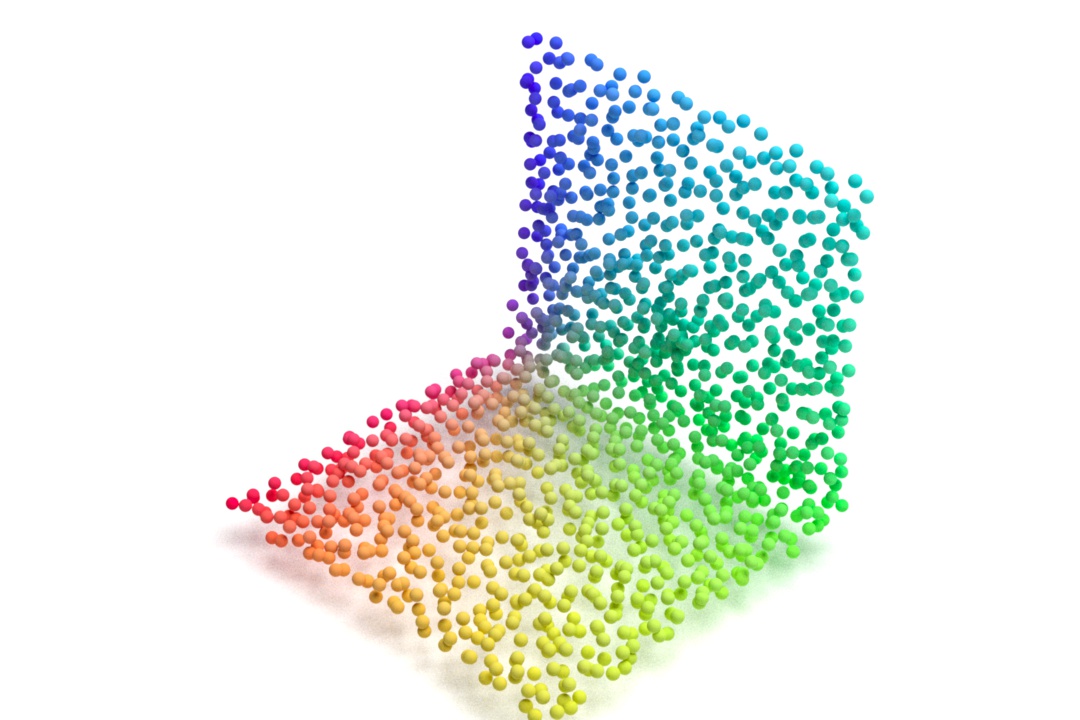}};
    \spy on (-.7,-0.4) in node [left] at (-.3,.7);
  \end{tikzpicture}
}
\subfloat{ 
  \begin{tikzpicture}[spy using outlines={circle,red,magnification=3,size=1.5cm, connect spies}]
    \node {\includegraphics[width=.16\textwidth]{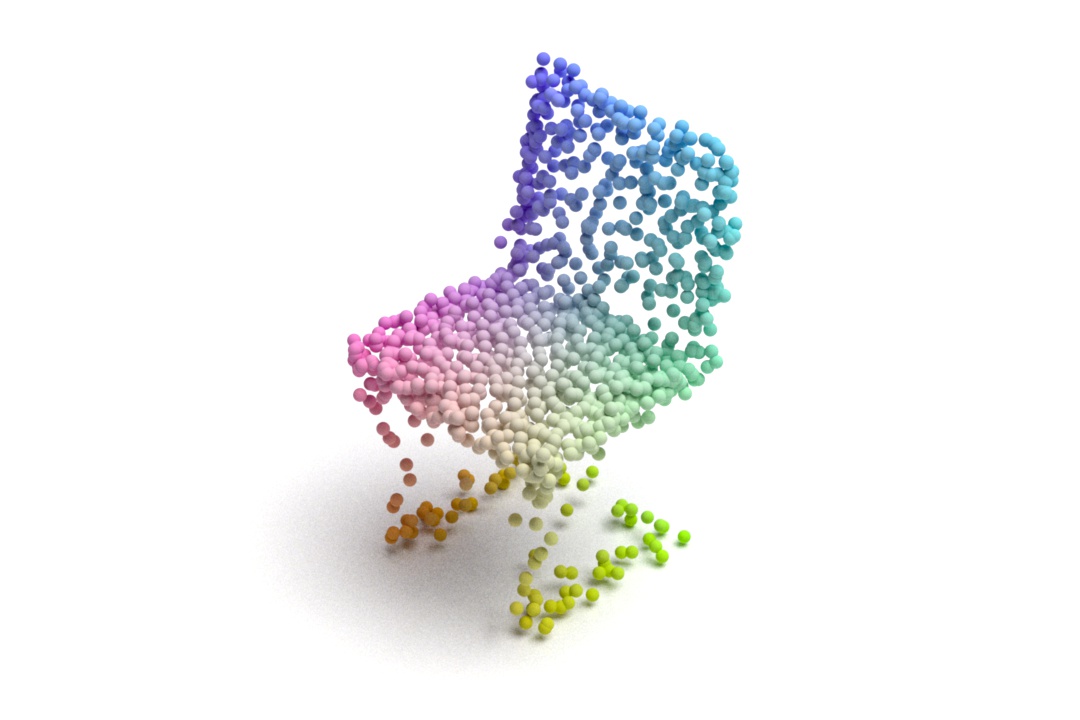}};
    \spy on (.0,-0.65) in node [left] at (-.3,.7);
  \end{tikzpicture}
}
\subfloat{ 
  \begin{tikzpicture}[spy using outlines={circle,red,magnification=3,size=1.5cm, connect spies}]
    \node {\includegraphics[width=.16\textwidth]{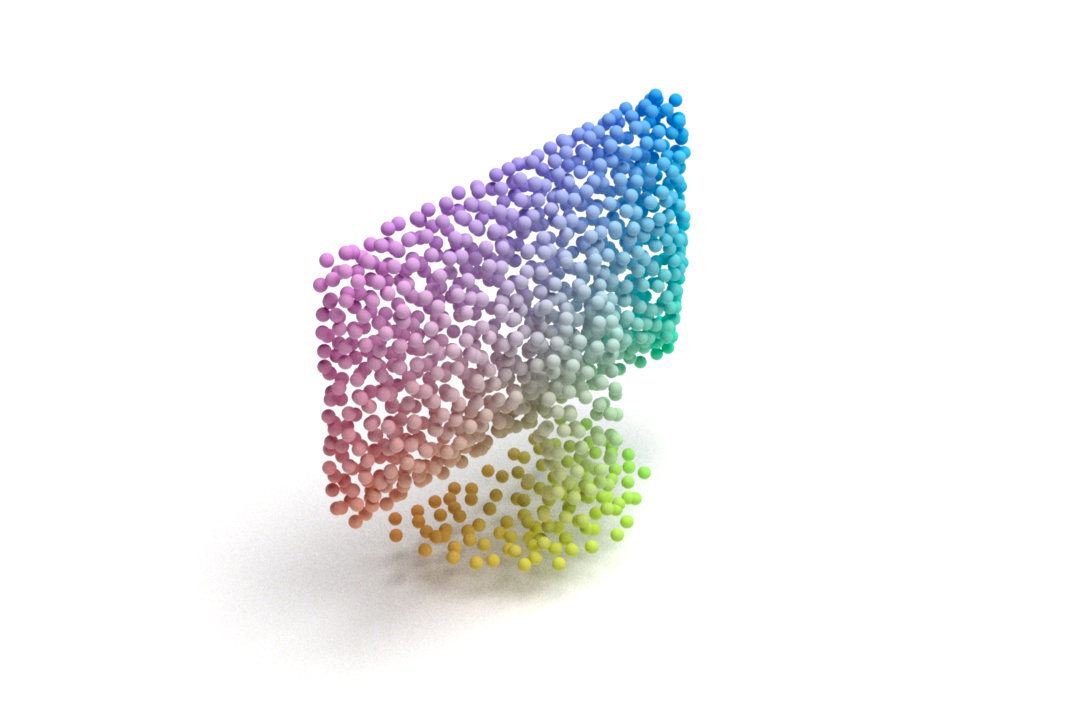}};
    \spy on (.08,-0.4) in node [left] at (-.3,.7);
  \end{tikzpicture}
}
\\
\subfloat{ 
\begin{tikzpicture}[spy using outlines={circle,red,magnification=3,size=1.5cm, connect spies}]
    \node {\includegraphics[width=.16\textwidth]{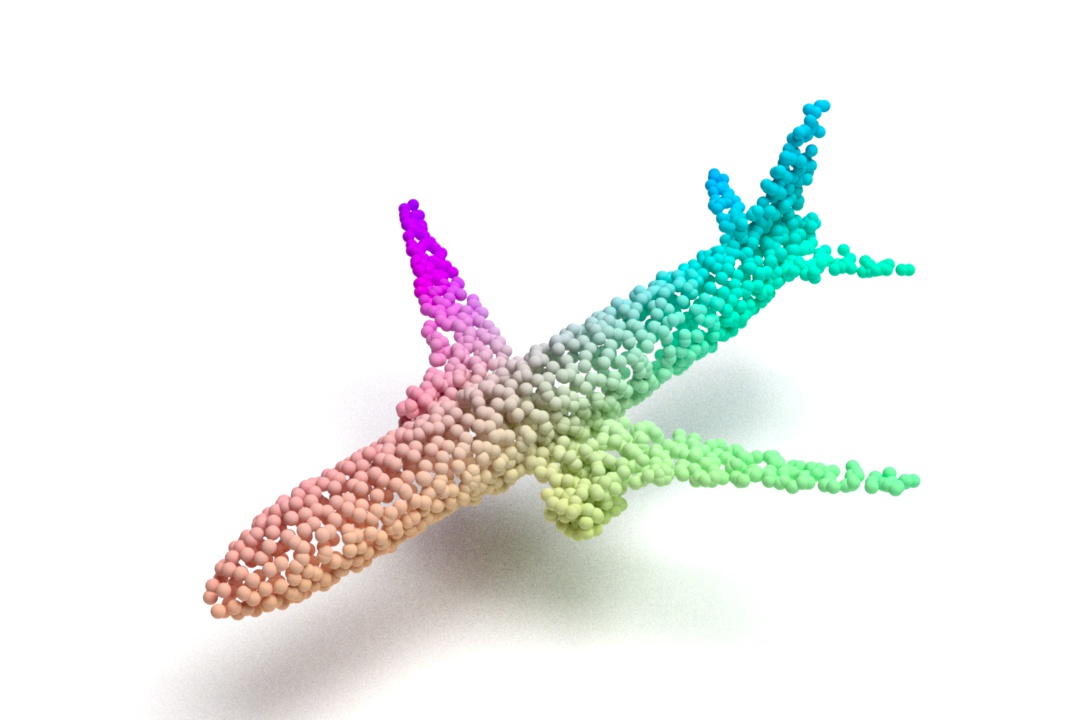}};
    \spy on (.07,-0.35) in node [left] at (-.3,.7);
\end{tikzpicture}

}
\subfloat{ 
  \begin{tikzpicture}[spy using outlines={circle,red,magnification=3,size=1.5cm, connect spies}]
    \node {\includegraphics[width=.16\textwidth]{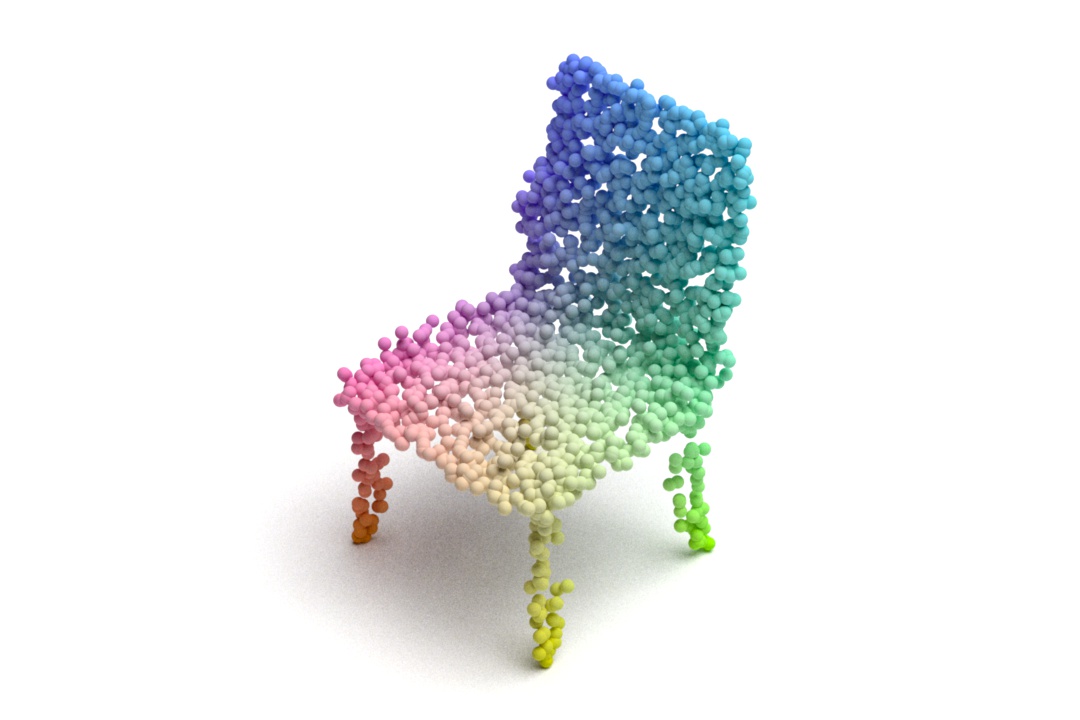}};
    \spy on (.02,-0.4) in node [left] at (-.3,.7);
\end{tikzpicture}
}
\subfloat{ 
  \begin{tikzpicture}[spy using outlines={circle,red,magnification=3,size=1.5cm, connect spies}]
    \node {\includegraphics[width=.16\textwidth]{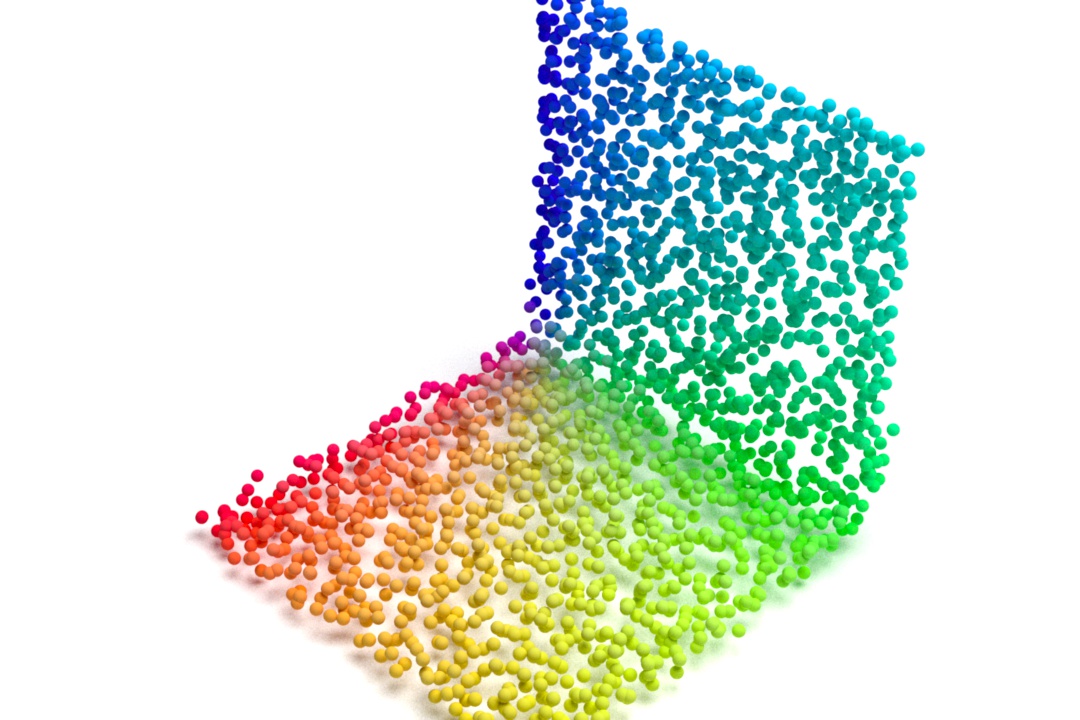}};
    \spy on (-.8,-0.45) in node [left] at (-.3,.7);
  \end{tikzpicture}
}
\subfloat{ 
  \begin{tikzpicture}[spy using outlines={circle,red,magnification=3,size=1.5cm, connect spies}]
    \node {\includegraphics[width=.16\textwidth]{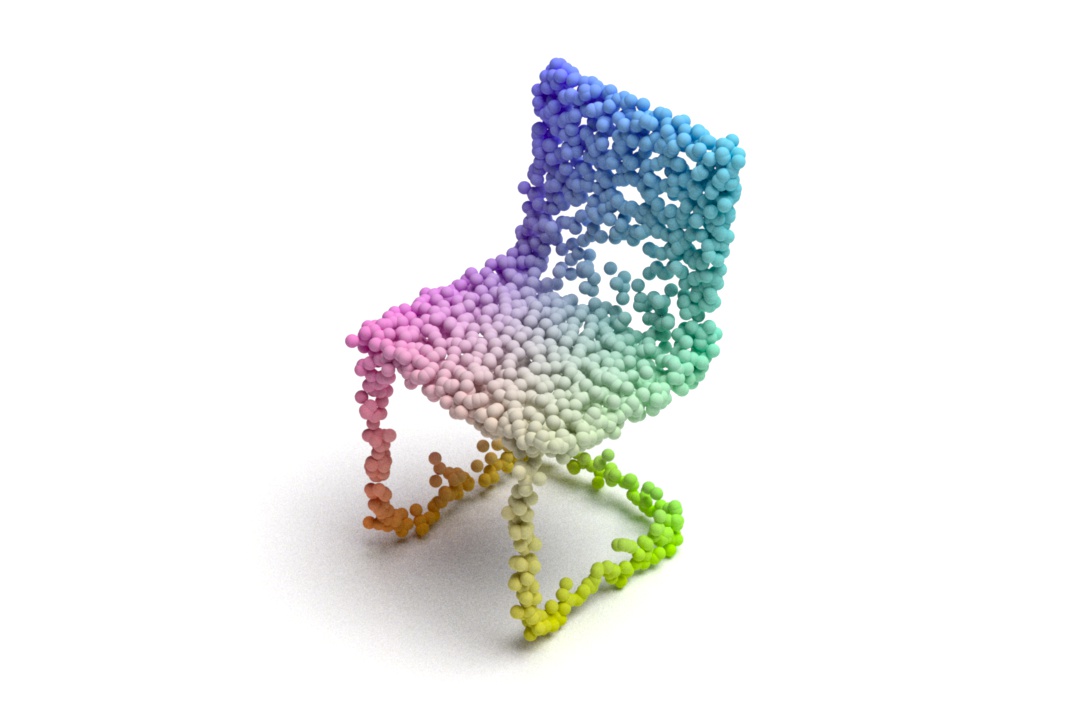}};
    \spy on (.0,-0.65) in node [left] at (-.3,.7);
  \end{tikzpicture}
}
\subfloat{ 
  \begin{tikzpicture}[spy using outlines={circle,red,magnification=3,size=1.5cm, connect spies}]
    \node {\includegraphics[width=.16\textwidth]{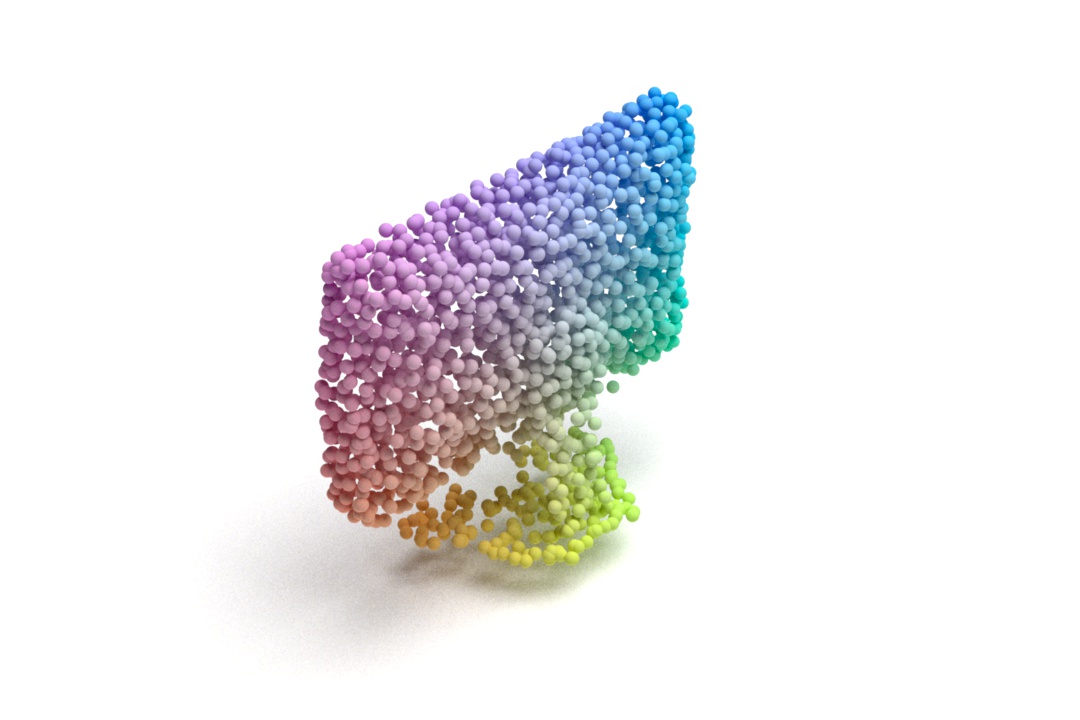}};
    \spy on (.05,-0.45) in node [left] at (-.3,.7);
  \end{tikzpicture}
}
\caption{The reconstruction results for a subset of objects from the ShapeNet
\cite{chang2015shapenet} dataset. The first row shows the ground-truth point
clouds. The second and third rows display the reconstructed point clouds using
the chamfer distance and earth mover's distance, respectively.} 
\label{fig:qualitative_reconstruction}
\end{figure*}

\begin{figure} 
\centering
\begin{tikzpicture} 
  \begin{axis}[xlabel=Bitrate,ylabel=Classification Accuracy] 
  \addplot[dashed,mark=*,color=black] plot coordinates { (0.0, 0.88442) (1.676785, 0.88442) }; 
  \addlegendentry{$Uncompressed$}
  \addplot[smooth,mark=x,color=purple] plot coordinates { (0.036987, 0.835526) (0.276215, 0.861842) (1.133270, 0.828947) (1.443545, 0.822368) };
  \addlegendentry{$CD$} 
  \addplot[smooth,mark=*,color=cyan] plot coordinates { (0.069773, 0.808442) (0.552924, 0.829545) (1.126253, 0.845373) (1.676785, 0.834821) }; 
  \addlegendentry{$EMD$}
  \end{axis} 
\end{tikzpicture} 
\caption{A comparison of the classification accuracy for uncompressed and
decompressed point cloud data using the chamfer distance (CD) and earth mover's
distance (EMD).} 
\label{fig:accuracy_comparison}
\end{figure}
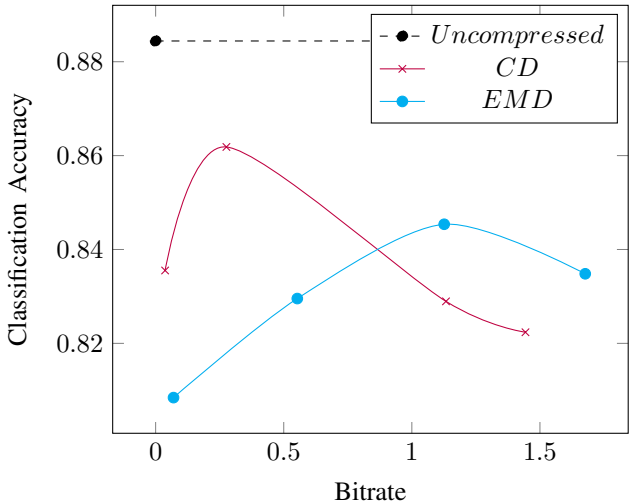

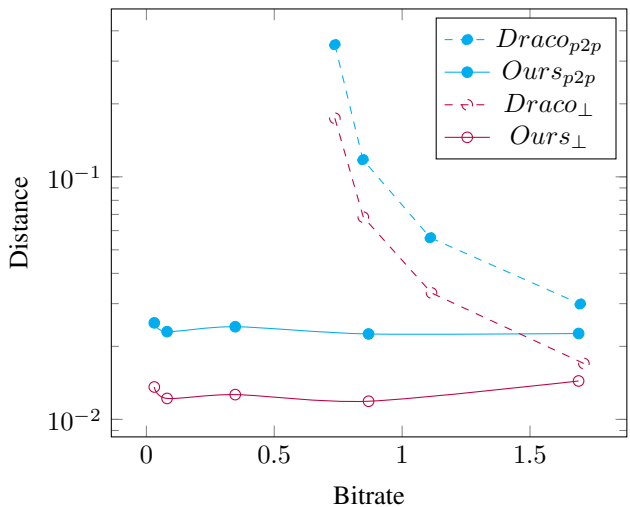
\begin{figure} 
\centering
\begin{tikzpicture} 
  \begin{axis}[xlabel=Bitrate,ylabel=Distance,ymode=log] 
    \addplot[dashed,mark=*,color=cyan] plot coordinates { (0.7368, 0.3509) (0.8468, 0.1177)(1.11034, 0.0561) (1.6965, 0.0299)};
    \addlegendentry{$Draco_{p2p}$} 
    \addplot[smooth,mark=*,color=cyan] plot coordinates { (0.0309,0.025) (0.0808, 0.023) (0.3474,0.0241 ) (0.8687,0.0225 )(1.690192, 0.0226) };
    \addlegendentry{$Ours_{p2p}$} 
    \addplot[dashed,mark=o,color=purple] plot coordinates { (0.737, 0.1744) (0.8483, 0.0682) (1.115, 0.0332) (1.7107, 0.0170)}; 
    \addlegendentry{$Draco_\perp$} 
    \addplot[smooth,mark=o,color=purple] plot coordinates { (0.0309, 0.0136) (0.0808,  0.0122) (0.3474, 0.01265 ) (0.8687, 0.01188)(1.690, 0.0144 ) };
    \addlegendentry{$Ours_\perp$} 
  \end{axis} 
\end{tikzpicture} 
\caption{The reconstruction quality of objects from the ShapeNet
\cite{chang2015shapenet} dataset in terms of the point-to-point (p2p) and
point-to-surface ($p_\perp$) distances at different bitrates.} 
\label{fig:draco_comparison}
\end{figure}

\begin{figure}
\centering 
\begin{tikzpicture} 
\begin{axis}[ xlabel=Latent Vector Size, ylabel=Chamfer Distance]
  \addplot[mark=*,cyan] plot coordinates {(1024,0.3184) (1020,0.3186)
  (1024-16,0.3438) (1024-32,0.3437) (1024-64,0.3515) (1024-128,0.3554)
  (1024-256,0.3815) (1024-300,0.3878) (1024-325,0.397) (1024-340,0.402)
};
\end{axis} 
\end{tikzpicture} 
\caption{The increase in reconstruction loss using a truncated latent vector to
enable variable bitrates.} 
\label{fig:variable_bitrate} 
\end{figure}

\subsection{Reconstruction Evaluation} 
Measuring the reconstruction capability of our model is crucial to evaluating
its performance. However, direct quantitative comparisons between two point
clouds is complicated by the property that the set of points comprising a point
cloud is invariant to permutations. To address this issue, we utilized the
chamfer distance (CD) \cite{barrow1977parametric} and earth mover's distance
(EMD) \cite{rubner2000earth}. The CD and EMD are two permutation-invariant
metrics for comparing sets of points introduced in \cite{fan2017point}. 

Initially, we used the same metrics (CD and EMD) to calculate the reconstruction
loss and to train our model. Table~\ref{tab:loss_per_category} establishes a
benchmark on the ShapeNet dataset using the CD, EMD, and \textit{F}-score
metrics. Despite the fact that both of the losses are differentiable almost
everywhere, the CD is more efficient to compute in comparison to the EMD. In our
experiments, we found that the training process converges faster when using the
CD compared to the EMD. Notwithstanding, the EMD metric produces slightly
higher-fidelity qualitative reconstruction results as shown in
Fig.~\ref{fig:qualitative_reconstruction}.

In a second experiment, we measured the performance of our network against Draco
1.4.1 \cite{draco}, a state-of-the-art open-source library for compressing and
decompressing 3D geometric meshes and point clouds. For this experiment, we
again used ShapeNet and trained our model using the CD loss. The comparison was
done using the point-to-point and point-to-surface distance metrics. These
metrics are widely used to quantify 3D surface reconstruction quality and are
described in \cite{tian2017geometric}. Fig.~\ref{fig:draco_comparison} shows
that our architecture outperforms Draco, especially at lower bitrates. 


\subsection{Task-Specific Evaluation} 
An important and practical use case of compressed point cloud data is to store
large amounts of 3D models and then preprocess them later on for downstream
functions. To quantify the reconstruction quality of such tasks, we used a
PointNet \cite{qi2017pointnet} model to  classify point cloud objects from
ModelNet40. In this experiment, we trained a model using uncompressed data and
determined its initial accuracy. Then, we used the trained model to classify the
decompressed data and examined its accuracy against the original data. Note
that both the uncompressed and decompressed data were compared using the same
metric (e.g., CD or EMD).

Fig.~\ref{fig:accuracy_comparison} shows that our compression method preserves
the semantic structure of the objects, even at low bitrates, while training with
different loss functions. We can achieve similar accuracy with the CD metric on
the compressed data (86\%) when compared against the actual data (88\%). An
interesting observation from this experiment is that the classification accuracy
increases gradually with the bitrate, up to a certain optimal point, and then
decreases. Based on these results, we hypothesize that the bitrate constraint
acts as a regularizer that helps the model to better generalize.

\subsection{Variable Bitrate Evaluation} 
This experiment showcases the possibility of using a single trained model to
compress point cloud data at varying bitrates. At training time, the weighted
entropy loss function \eqref{eq:weighted_entropy} penalizes the
lower-indexed vector elements more during the compression process.
Consequently, this ensures that the high-frequency information is carried
through the higher-indexed vector elements while the lower-indexed vector
elements carry the overall shape information. The exponential decaying function
\eqref{eqn:weight_decay} is used for assigning a weight to each vector element
where $\omega_i$ is the $i$-th element in the weight vector. 

For this experiment our model was trained using the CD metric, and the values 15
and 0.003 for variables $a$ and $b$, respectively, were used for
\eqref{eqn:weight_decay}. At test time, we truncate the latent vector according
to the bitrate setting and only compress the truncated vector instead of the
whole vector. In Fig.~\ref{fig:variable_bitrate}, we illustrate the relationship
between the loss and the number of vector elements used in the reconstruction.
As we truncate the higher-indexed vector elements the distance between the
actual and compressed point cloud gradually increases. This naturally results in
a less accurate reconstruction. 




\section{Conclusion and Future Work} 
\label{sec:conclusion_and_future_work}
We presented a deep learning-based approach that is capable of variable rate raw
point cloud compression. Variable-rate compression is enabled by a tunable loss
function that distributes information non-uniformly among the latent vector
elements. This method alleviates the need for the training and storage of
multiple models over a range of bitrates. The efficacy of our model was shown
through the analysis of different components and the presentation of an
extensive evaluation on the individual impact of these factors. What's more, we
established new benchmarks on publicly available datasets and we showed that our
model surpasses state-of-the-art open-source software for compressing and
decompressing 3D point clouds. In future work, we will investigate extending the
range of bitrates a single model can support. 

\section*{Acknowledgments}
The authors acknowledge the Texas Advanced Computing Center (TACC) at the
University of Texas at Austin for providing software, computational, and storage
resources that have contributed to the research results reported within this
paper.

\newpage
\bibliographystyle{IEEEtran}
\bibliography{IEEEabrv,variable_rate_compression_for_raw_3d_point_clouds}   
\end{document}